\documentclass[sigconf,numbers,sort&compress]{acmart}
\AtBeginDocument{%
  }

\usepackage{amsmath,amsfonts}
\usepackage{algorithmic}
\usepackage{graphicx}
\usepackage{textcomp}
\usepackage{xcolor}
\usepackage{hhline}
\usepackage{pifont}
\usepackage{svg}
\usepackage{multirow}
\usepackage{datetime}
\usepackage{tabularx}
\usepackage{threeparttable}
\usepackage{booktabs}
\usepackage{makecell}
\usepackage{float}
\usepackage{subcaption}
\usepackage{hyperref}

\usepackage{wrapfig}
\usepackage{tikz}
\usetikzlibrary{shapes,arrows}
\usepackage{pgfplots}
\pgfplotsset{compat=1.9}
\usepackage{mathtools}

\usepackage{siunitx}
\usepackage{tikz}
\usetikzlibrary{shapes, arrows, decorations, positioning, calc, shadows, trees, mindmap}
\usepackage{listings}
\usepackage[figuresright]{rotating}
\usepackage{afterpage}
\usepackage{comment}
\usepackage{lineno}

\usepackage[nolist]{acronym}

\usepackage[frozencache,cachedir=.]{minted}
\usepackage{listings}

\begin{acronym}[MACHU]
  \acro{m3kg}[MRM3]{Machine Readable ML Model Metadata}
  \acro{ml}[ML]{machine learning}
  \acro{ai}[AI]{artificial intelligence}
  \acro{mlops}[MLOps]{Machine Learning Operations}
  \acro{mse}[MSE]{mean squared error}
  \acro{mpe}[MPE]{mean percentage error}
  \acro{kg}[KG]{knowledge graph}
  \acro{json}{JavaScript Object Notation}
  \acro{flops}[FLOPs]{Floating Point Operations}
  \acro{hf}[HF]{Hugging Face}
  \acro{gnn}[GNN]{graph neural networks}
  \acro{llm}[LLM]{large language models}
  \acro{rag}[RAG]{Retrieval Augmented Generation}
\end{acronym}

\copyrightyear{2025} 
\acmYear{2025} 
\setcopyright{acmlicensed}\acmConference[NetAISys'25 ]{3rd International Workshop on Networked AI Systems}{June 27, 2025}{Anaheim, California, United States}
\acmBooktitle{3rd International Workshop on Networked AI Systems (NetAISys '25 ), June 27, 2025, Anaheim, California, United States}
\acmDOI{10.1145/3711875.3736685}
\acmISBN{979-8-4007-1453-5/25/06}

\begin{document}

\title{MRM3: Machine Readable ML Model Metadata}

\author{Andrej \v{C}op, Bla\v{z} Bertalani\v{c}, Marko Grobelnik, Carolina Fortuna}
\affiliation{%
  \institution{Jo\v{z}ef Stefan Institute}
  \city{Ljubljana}
  \country{Slovenia}
}
\email{ac8560@student.uni-lj.si}

\renewcommand{\shortauthors}{A. Čop et al.}

\begin{abstract}
As the complexity and number of \ac{ml} models grows, well-documented \ac{ml} models are essential for developers and companies to use or adapt them to their specific use cases.
Model metadata, already present in unstructured format as model cards in online repositories such as Hugging Face, could be more structured and machine readable while also incorporating  environmental impact  metrics such as energy consumption and carbon footprint. 
Our work extends the existing State of the Art by defining a structured schema for ML model metadata focusing on machine-readable format and support for integration into a \ac{kg} for better organization and querying, enabling a wider set of use cases.
Furthermore, we present an example wireless localization model metadata dataset consisting of 22 models trained on 4 datasets, integrated into a Neo4j-based \ac{kg} with 113 nodes and 199 relations.
\end{abstract}





\maketitle

\section{Introduction}
\ac{ml} models have become essential for modern intelligent systems, enabling automation and data-driven decision making across numerous domains such as modern 5G/6G  networks~\cite{Letaief2019TheNetworks} or Internet of Things~\cite{Cui2018AThings}. 
As the complexity and diversity of ML models grow, systematic representation and organization of model and dataset metadata become increasingly important. 
Model metadata containing  information describing a model's development context and characteristics such as  training process, performance, bias, and input datasets is the foundation for informed decisions regarding model selection, deployment, retraining, and optimization~\cite{pepe_hugging}. 

Model repositories such as \ac{hf}\footnote{\url{https://huggingface.co/}}, focus on providing model and dataset cards, which consist of unstructured metadata intended primarily for human readability~\cite{analysis_ai_model_cards}. 
Model and dataset cards consists of some standardized representation, and existing work attempts to define what to include in the documentation~\cite{Mitchell2019ModelReporting}. 
Specifically, for ML dataset cards, there is existing work on machine readability. 
Structured representation and standard format of dataset cards allow not only human readability but also enable intelligent searching for the appropriate datasets by automated means, therefore accelerating the creation of new models on existing, well-documented datasets~\cite{Giner-Miguelez2023ADatasets, describeml, Croissant_paper}.

Current approaches to model documentation, such as model cards, provide valuable descriptive information but typically exist as unstructured text, making systematic comparison and selection challenging. 
Although significant progress has been made in dataset documentation through frameworks for ML datasets, such as DescribeML~\cite{describeml} and Croissant~\cite{Croissant_paper}, equivalent structured representations for model metadata remain underdeveloped.
The absence of a standardized machine-readable representation of model metadata inhibits the development of fully automated intelligent systems for model orchestration and selection. 

Sustainability metrics such as carbon footprint and energy consumption are becoming important as the environmental impact of ML model training and inference is explored~\cite{wu2022sustainable}.
These metrics should be consistently included in such documentation allowing direct comparison between different models and enabling the selection of the best ML model based on multiple criteria, such as performance metrics, computational requirements, and environmental impact, which are specifically useful in low-powered network edge devices.
To achieve this, a novel approach to representation of structured model metadata is needed. 
A graph database that connects different entities with relations, improves the ability to select related ML models based on various criteria, as different ML models can be highly correlated.
For example, models trained on the same dataset with different architecture or parameters or the models for same service. 
This paper addresses these limitations by introducing a solution for representing ML model metadata through a \ac{kg} approach. 
Our contributions are as follows.

\begin{itemize}
    \item We identify a taxonomy of ML model metadata and propose an ontology defining the relationships between model metadata entities, enabling structured representation in a \ac{kg}.
    \item Based on proposed ontology we define a Neo4j based \textit{\ac{m3kg}} knowledge graph. We show an example \ac{kg} using model metadata from 22 trained wireless localization models. The localization \ac{m3kg} consists of 113 nodes and 199 relations and enables querying of models by lowest environmental impact while satisfying the required accuracy. 
    \item  By open-sourcing our code, providing an example localization model metadata dataset, and offering a standardized \ac{json} schema for collecting and representing model metadata, we enable and simplify the implementation of the proposed \ac{m3kg} for any user or organization.
\end{itemize}

The remainder of this paper is organized as follows. In Section~\ref{sec:related-work} we discuss the related work, while in Section~\ref{sec:taxonomy-ontology} we identify ML model metadata taxonomy and propose ontology for \ac{m3kg}.
In Section~\ref{sec:dataset-kg} we present an example localization \ac{m3kg}. 
Finally, we conclude the paper in Section~\ref{sec:conclusion}.

\section{Related Work}
\label{sec:related-work}

Existing work on \ac{ml} dataset and model documentation has highlighted the need for more structured and precise metadata representation. 
Giner-Miguelez et al.~\cite{Giner-Miguelez2023ADatasets,describeml} proposed a domain-specific language (DSL) and a DescribeML tool to document ML dataset structure, provenance, and fairness.
On the other hand, Akhtar et al.~\cite{Croissant_paper} introduced Croissant, a metadata format adopted by \ac{hf}, improving dataset discoverability and interoperability across the ML field. 
While we do not focus on structured dataset documentation this related work inspired machine readability of our proposed model metadata structure.

Pepe et al.~\cite{pepe_hugging} and Liang et al.\cite{analysis_ai_model_cards} analyzed \ac{hf} model cards, finding limited disclosure of datasets, biases, and licenses, with Liang et al. noting improved downloads with more detailed cards.
Similarly, Yang et al.~\cite{yang2024navigatingdatasetdocumentationsai} performed an analysis of dataset cards on \ac{hf} platform. 
Model and dataset cards on \ac{hf} are in a human-readable format and provide useful documentation for developers but lack a consistent schema and ontology, which we address in this work.

Mitchell et al.~\cite{Mitchell2019ModelReporting} proposed model cards for transparent reporting of ML model performance across diverse conditions at Google. Beyond the research paper, Google has developed a suite of tools for structured model documentation, including the Model Card Toolkit~\cite{ModelCardToolkit}, which provides a \ac{json} schema for standardizing model reporting. 
Their ML Metadata (MLMD) library further enables tracking and managing model metadata throughout the ML lifecycle~\cite{MLMetadata}.
While Google's \ac{json} schema standardizes model reporting, our work extends this by defining a structured schema for ML model metadata focusing on machine-readable format and support for integration into a \ac{kg} for better organization and querying, enabling a wider set of use cases.

McMillan-Major et al.~\cite{McMillan-Major2021ReusableCards} developed reusable natural language processing documentation templates for \ac{hf}. 
On the other hand, Bhat et al.~\cite{Bhat2023AspirationsTraceability} and Crisan et al.~\cite{Crisan2022InteractiveDocumentation} addressed documentation gaps, introducing DocML for ethical reporting and interactive model cards to aid non-expert understanding, respectively.
Amith et al.~\cite{Amith2022TowardReport} formalized model cards into an OWL2 ontology for machine-readable bioinformatics models, while Ajibode et al.~\cite{ajibode2025semanticversioningopenpretrained} critiqued inconsistent versioning in \ac{hf} language models, recommending semantic standards.

In contrast to these efforts on dataset and model documentation, our work uniquely integrates a taxonomy, ontology, and Neo4j-based \ac{kg} for ML model metadata.

\section{ML Model Metadata Taxonomy and Ontology}
\label{sec:taxonomy-ontology}
Documentation of datasets and \ac{ml} models enables users to understand and effectively utilize published open-sourced \ac{ml} models in their applications. 
An example of such documentation is presented as dataset and model cards on a popular model repository platform \ac{hf}.
Based on analyses of existing work and proposed approaches to \ac{ml} model documentation we can identify a common structure and taxonomy for such documentation, focusing on attributes that are important for the automated selection of \ac{ml} models.  

In this section, we define the taxonomy of \ac{ml} model metadata and using an established methodology for \ac{kg} development~\cite{Tamasauskaite2023DefiningReview} we can construct an ontology, and develop a \ac{kg} for \ac{m3kg}.

\subsection{Identified Taxonomy}
\label{subsec:taxonomy}

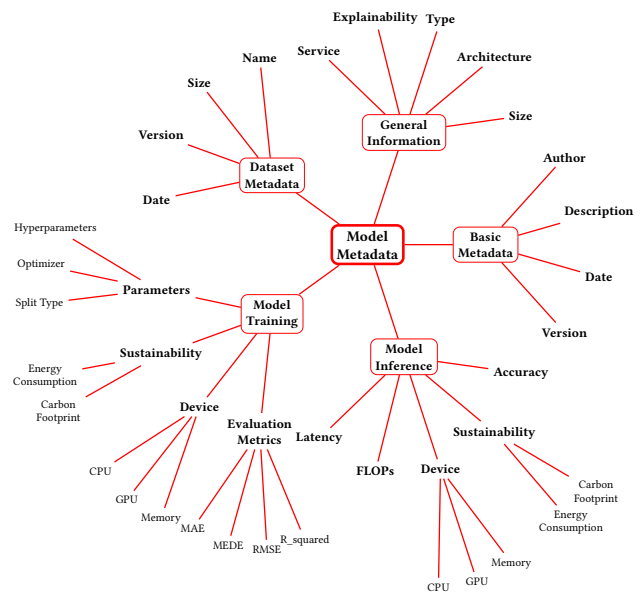
\begin{figure}[htbp]
\centering
\resizebox{1\linewidth}{!}{
\begin{tikzpicture}[
    scale=1,
    grow cyclic,
    edge from parent/.style = {draw, -, thick, red},
    sloped,
    box/.style = { shape=rectangle, rounded corners, draw=red },
    level/.style = { align=center },
    level 0/.style = { level, font=\bfseries\normalsize, align=center, ultra thick },
    level 1/.style = { level, level distance=8em, font=\bfseries\small, sibling angle=72 },
    level 2/.style = { level, level distance=8em, font=\bfseries\small, sibling angle=32 },
    level 3/.style = { level, level distance=8em, font=\footnotesize, sibling angle=19 },
    level 4/.style = { level, level distance=7em, font=\scriptsize, sibling angle=12 }
]
\node [level 0, box] {Model\\Metadata}
    child [level 1] { node [box] {Model\\Training}
        child [level 2] { node {Parameters}
            child [level 3] { node {Hyperparameters} }
            child [level 3] { node {Optimizer} }
            child [level 3] { node {Split Type} }
        }
        child [level 2] { node {Sustainability}
            child [level 3] { node {Energy\\Consumption}
            }
            child [level 3] { node {Carbon\\Footprint} }
        }
        child [level 2] { node {Device}
            child [level 3] { node {CPU} }
            child [level 3] { node {GPU} }
            child [level 3] { node {Memory} }
        }
        child [level 2] { node {Evaluation\\Metrics}
            child [level 3] { node {MAE} }
            child [level 3] { node {MEDE} }
            child [level 3] { node {RMSE} }
            child [level 3] { node {R\_squared} }
        }
    }
    child [level 1] { node [box] {Model\\Inference}
        child [level 2] { node {Latency} }
        child [level 2] { node {FLOPs} }
        child [level 2] { node {Device}
            child [level 3] { node {CPU} }
            child [level 3] { node {GPU} }
            child [level 3] { node {Memory}
            }
        }
        child [level 2] { node {Sustainability}
            child [level 3] { node {Energy\\Consumption}
            }
            child [level 3] { node {Carbon\\Footprint} }
        }
        child [level 2] { node {Accuracy} }
    }
    child [level 1] { node [box] {Basic\\Metadata}
        child [level 2] { node {Version} }
        child [level 2] { node {Date} }
        child [level 2] { node {Description} }
        child [level 2] { node {Author} }
    }
    child [level 1] { node [box] {General\\Information}
        child [level 2] { node {Size} }
        child [level 2] { node {Architecture} }
        child [level 2] { node {Type} }
        child [level 2] { node {Explainability} }
        child [level 2] { node {Service} }
    }
    child [level 1] { node [box] {Dataset\\Metadata}
    child [level 2] { node {Name} }
    child [level 2] { node {Size} }
    child [level 2] { node {Version} }
    child [level 2] { node {Date} }
    }
;
\end{tikzpicture}
}
\caption{Taxonomy of ML model metadata.}
\label{fig:taxonomy}
\end{figure}

Based on prior work on model card structure~\cite{Mitchell2019ModelReporting, Amith2022TowardReport}, analysis of model cards on \ac{hf} model repository~\cite{analysis_ai_model_cards, McMillan-Major2021ReusableCards} and inspired by the previously proposed structure of machine-readable dataset metadata~\cite{Croissant_paper, describeml, Giner-Miguelez2023ADatasets} we identify taxonomy of \ac{ml} model metadata. 
This taxonomy provides a structure on what specific metadata to extract when training and testing \ac{ml} models. The identified taxonomy is presented in Fig.~\ref{fig:taxonomy} and further explained in subsequent paragraphs.

The main branches of model metadata are model training, model inference, dataset metadata, general information and basic metadata as depicted in Fig.~\ref{fig:taxonomy}. 

Model training metadata refers to information about how and where the model was trained and which evaluation metrics are collected. 
For example, model training parameters and hyperparameters and evaluation metrics such as root mean squared error (RMSE) or mean absolute error (MAE). However, it is not limited to metrics for regression but can be extended to include evaluation metrics for tasks such as clustering and classification.
Further, device metadata describes the CPU, GPU, and memory information of the device on which the model was trained.
We identified sustainability as an important part of model training~\cite{wu2022sustainable}.
Sustainability is further divided into energy consumption and carbon footprint as can be seen in the left bottom part of Fig.~\ref{fig:taxonomy}. 

Model inference metadata, similarly to training metadata, includes sustainability metrics and information about the device on which the inference was performed.
It also provides end-to-end latency information for inference requests and computational complexity which refers to the number of \ac{flops} required for running inference on a single input sample~\cite{Desislavov2023TrendsLearning}. 

Dataset metadata, presented as \textit{Dataset} in Fig.~\ref{fig:taxonomy}, includes essential information to uniquely identify datasets such as date, version, size and name, following the commonly found dataset metadata proposed in DescribeML framework~\cite{describeml} and Croissant metadata format~\cite{Croissant_paper}.

Similarly, general model information and basic metadata include information about date, version, and size as well as model architecture, the service for which the model was intended, and the architecture's explainability. 
It can also include information such as a description of the models and a list of authors. 
General model information is depicted in the top right of the ML model metadata taxonomy and basic metadata on the right side in Fig.~\ref{fig:taxonomy}.

While information such as license, citation details, and associated research papers are common in model cards~\cite{Mitchell2019ModelReporting, McMillan-Major2021ReusableCards} we don't identify them as important in our model metadata taxonomy as they are not crucial for use cases such as automated selection of ML models. However, the taxonomy can be extended to also include such information.

\subsection{Ontology}
\label{subsec:ontology}

Taxonomy of ML model metadata, identified in Section~\ref{subsec:taxonomy}, helps developers decide on which metadata to collect and store during the process of development, training, and testing of ML models, however, this metadata by itself does not support connecting different ML models and creating relations between them. 
For this reason, we define ontology~\cite{Guarino2009WhatOntology} of ML model metadata. An ontology defines the structure of the \ac{kg} and allows a model of how the \ac{kg} is represented in a structured way with relations between concepts~\cite{Tamasauskaite2023DefiningReview}.

\begin{figure}[ht]
    \centering
    \includegraphics[width=1\linewidth]{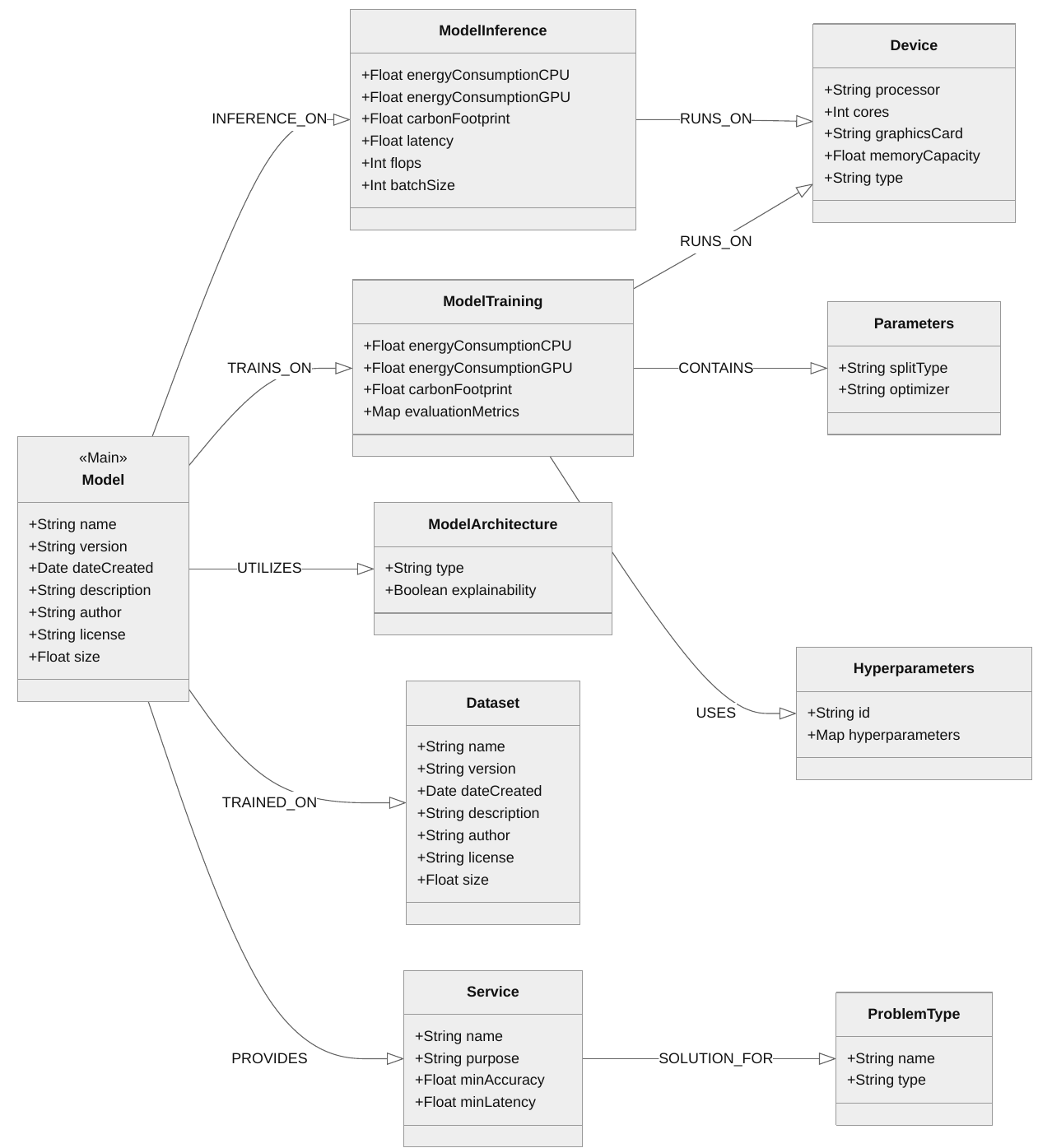}
    \caption{Ontology of ML model metadata.}
    \label{fig:ontology}
\end{figure}

Using the identified taxonomy in Section~\ref{subsec:taxonomy} and presented in Fig.~\ref{fig:taxonomy}, we propose an ontology of ML model metadata presented in Fig.~\ref{fig:ontology}.
The ontology is proposed based on relations between definitions in the taxonomy and using the methodology for creating knowledge graphs~\cite{Tamasauskaite2023DefiningReview}.

As we can see from Fig.~\ref{fig:ontology}, general information about the model is contained in the main node, called \textit{Model}.
A relation \textit{TRAINED\_ON} connecting node \textit{Model} to node \textit{Dataset} defines the dataset that the model was trained on including dataset metadata. 
The relation \textit{PROVIDES} connects models to their respective service and its requirements for example localization service, text generation, and image classification. 
\textit{Service} node connects to \textit{ProblemType} with relation \textit{SOLUTION\_FOR}, which defines the type of ML problem such as regression, classification, clustering etc. 
The relation \textit{UTILIZES} and node \textit{ModelArchitecture} define the architecture of the ML model.

In Fig.~\ref{fig:ontology} we can see relation \textit{TRAINS\_ON} which connects node \textit{Model} to node \textit{ModelTraining}. 
\textit{ModelTraining} includes attributes concerning training of the ML model and itself connects to node \textit{Parameters}, and \textit{Hyperparameters} which define a set of parameters and hyperparameters that the model is trained with. 
Similarly, node \textit{ModelInference} describes attributes concerning ML models in prediction mode, such as latency and \ac{flops}.
As we can see in the bottom left of Fig.~\ref{fig:ontology}, both \textit{ModelTraining} and \textit{ModelInference} connect to \textit{Device} with relation \textit{RUNS\_ON} which specify the device and its hardware that the model was trained on or performed inference.

\subsection{Knowledge Graph}
\label{subsec:knowledge_graph}

Knowledge graphs are graph-structure knowledge bases~\cite{Tamasauskaite2023DefiningReview}. 
They use a graph data model to capture knowledge, providing an abstraction of domains, while capturing relations between entities in these domains. 
Ontologies define the structure of knowledge graphs, while queries enable clustering, summarization, and insight generation about the described domain~\cite{Hogan2021KnowledgeGraphs}.

To create a knowledge graph that can support use cases such as automated selection of ML models, a methodology of identifying data, constructing ontology, extracting knowledge, processing knowledge, and constructing the knowledge graph was used~\cite{Tamasauskaite2023DefiningReview}. Based on the constructed ontology in Section~\ref{subsec:ontology} and Fig.~\ref{fig:ontology} we can create \ac{kg} for any dataset of model metadata that follows the identified taxonomy of model metadata in Section~\ref{subsec:taxonomy} and Fig.~\ref{fig:taxonomy}. 

We utilize Neo4j Graph Database Platform\footnote{\url{https://neo4j.com/}} to support the construction of our knowledge graph and enable queries using Cypher language. 
Neo4j is one of the top open-source graph database~\cite{Guia2017GraphAnalysis} and is widely used in the knowledge graph development process~\cite{Tamasauskaite2023DefiningReview}.

\section{Localization MRM3}
\label{sec:dataset-kg}

In this section, we present an example \ac{m3kg} dataset and knowledge graph. 
The localization model metadata dataset\footnote{\url{https://doi.org/10.1145/3711875.3736685}} is based on model metadata collected from wireless localization models of different architectures and on different wireless datasets, namely \textit{Lumos5G}~\cite{Lumos5G}, \textit{LOG-a-TEC}~\cite{Bertalanič_Morano_Cerar_2022} (winter and spring), and \textit{UMU\footnote{The data used in this study was collected as part of the EU Nancy project and it will be publicly available soon.}}~\cite{gaia5g}, using low-code-localization pipeline.
Using identified taxonomy in Section~\ref{subsec:taxonomy} we collected the required metadata and inserted it into the Neo4j-based knowledge graph with the structure as defined in Section~\ref{subsec:ontology}.

\subsection{Json Schema}
\label{subsec:json_schema}

To have a consistent way of collection and structure of metadata we propose a model metadata \ac{json} schema. 
Which is available on \ac{m3kg} Github repository\footnote{\label{footnote:github}\url{https://github.com/sensorlab/MRM3}} and can be used to verify the structure of any \ac{json} ML model metadata dataset. 
It defines which metrics to collect, how to structure them in \ac{json} format.
The schema includes short descriptions of metrics and in which units of measurement they should be stored.

\subsection{Model Metadata Knowledge Graph}
\label{subsec:loc_kg}

From the localization training pipeline, we collected \ac{json} model metadata following the proposed \ac{json} schema in the preceding section for 4 different datasets and multiple architectures such as random forest, k-nearest neighbors and XGBoost. 
We developed a python interface to insert any \ac{json} file following the defined \ac{json} schema into a deployed Neo4j \ac{kg} instance\footnotemark[5].

Table~\ref{table:kg} presents the number of nodes and entities in our localization \ac{kg}. 
Our example \ac{kg} consists of 22 models trained on 4 datasets with models based on 4 architectures. 
All the training and inference was performed on one device and in total 44 nodes have relation \textit{RunsOn} to the device, as we can see from Table~\ref{table:kg}. 
In total it includes 113 nodes and 199 relations. 
The details of relations, nodes and which properties they contain are explained in Section~\ref{fig:ontology} and in Fig.~\ref{fig:ontology}.

\begin{table}[htb]
\centering
\begin{tabular}{p{0.5\linewidth}r}
\hline
\textbf{KG Entities} & \textbf{Quantity} \\
\hline
All Relations & 199 \\
All Nodes & 113 \\
Relation \textit{RunsOn} & 44 \\
Node \textit{Model} & 22 \\
Node \textit{Dataset} & 4 \\
Node \textit{ModelArchitecture} & 4 \\
Node \textit{Device} & 1 \\
\hline
\end{tabular}
\caption{Summary of localization \ac{m3kg}.}
\label{table:kg}
\end{table}

\subsection{Use Case}
\label{subsec:use_case}

Wireless localization \ac{m3kg} enables use cases such as the selection of ML models and providing insight into ML models and their metadata.
Specifically for network edge needs where efficient computation of model inference is crucial, \ac{m3kg} enables intelligent selection of ML models based on certain search criteria such as low computational complexity for inference or minimizing energy consumption. 

By integrating \ac{m3kg} to \ac{mlops} systems, which orchestrate the process of \ac{ml} training and deployment on distributed and edge infrastructure, this use case improves the process of orchestrating ML models and could improve energy consumption and inference latency on low power edge devices~\cite{COP2025104180}.

\subsection{Querying the Knowledge Graph}
\label{subsec:queryingKG}

\begin{figure}[ht]
    \centering
    \includegraphics[width=1\linewidth]{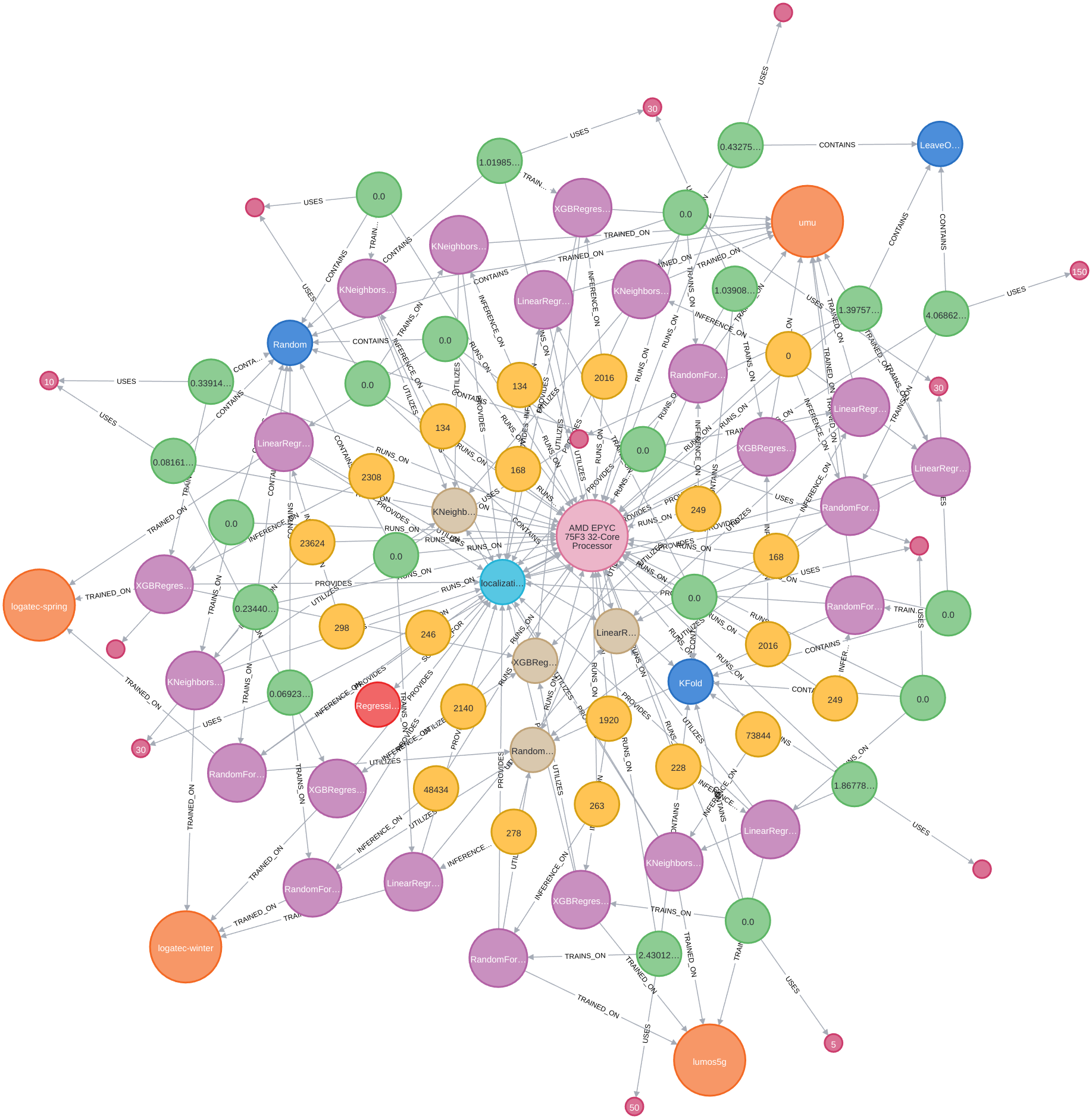}
    \caption{Localization \ac{m3kg} on Neo4j.}
    \label{fig:kg-graph}
\end{figure}

Cypher is a declarative language to query information from the Neo4j graph database~\cite{Guia2017GraphAnalysis}. 
It can be used to interact and query knowledge from any \ac{m3kg}. Using Cypher it is possible to query and display all data results in a graph depicted in Fig.~\ref{fig:kg-graph}. 
It displays all nodes and relations in the localization \ac{m3kg} with one property written in the center of each node. 
Each type of node is colored by different color, for example, dataset nodes are colored orange and there are four of them as we can confirm in Table~\ref{table:kg} summarizing the localization \ac{m3kg}.

To enable the use case described in Section~\ref{subsec:use_case} we provide an example Cypher query in Listing~\ref{listing:cypher}. 
The query returns models ordered by lowest energy consumption for model inference, while also providing computational complexity for inference and architecture of the model.

As we can see in Listing~\ref{listing:cypher}, it first matches \textit{Model} nodes that have associated \textit{Dataset} and \textit{ModelArchitecture} information in the \ac{kg}.
The query connects to \textit{ModelInference} nodes through the \textit{INFERENCE\_ON} relationships to extract inference metrics about energy consumption and computational complexity.
The results are ordered by ascending energy consumption values in the last line in Listing~\ref{listing:cypher}. 

\begin{listing}[ht]
\begin{small}
\begin{minted}[]{cypher}
MATCH (m:Model)-[:TRAINED_ON]->(d:Dataset)
MATCH (m)-[:UTILIZES]->(a:ModelArchitecture)
MATCH (i:ModelInference)-[:INFERENCE_ON]->(m)
RETURN m.name, 
       a.type as architecture,
       d.name as dataset,
       i.energyConsumption,
       i.flops
ORDER BY i.energyConsumption ASC
\end{minted}
\end{small}
\caption{Example code to query Neo4j based \ac{m3kg} using Cypher.}
\label{listing:cypher}
\end{listing}

The query output by Listing~\ref{listing:cypher} of the top five models is presented in Table~\ref{tab:model_performance}. The first column describes the model architecture, while the second specifies the dataset used for training. The third column lists the energy consumption per inference request (in joules), sorted in ascending order. The final column indicates the computational complexity measured in FLOPs.

\begin{table}[h]
\centering
\begin{tabular}{llrr}
\hline
\textbf{Architecture} & \textbf{Dataset} &\textbf{Energy (J)} & \textbf{FLOPs} \\
\hline
Random Forest & UMU & 0.072 & 249 \\
Random Forest & Lumos5G & 0.132 & 263 \\
XGBoost & LOG-a-TEC Winter & 0.284 & 140 \\
KNeighbors & UMU & 0.326 & 134 \\
Random Forest & LOG-a-TEC Spring & 0.370 & 246 \\
\hline
\end{tabular}
\caption{Output of example query in Listing~\ref{listing:cypher}.}
\label{tab:model_performance}
\end{table}

Executing the query in Listing~\ref{listing:cypher} takes approximately 3 ms on average, while retrieving all data shown in Fig.~\ref{fig:kg-graph} takes about 7 ms on average.

\subsection{Implementation Considerations}
\label{subsec:impl_cons}

To construct the model metadata \ac{kg}, the \ac{json} schema proposed in Section~\ref{subsec:json_schema} serves as a template to determine which data should be collected from the model training project, how it should be structured in \ac{json} files, and in which units of measurement the collected metrics should be stored.
Training code or training pipeline must be modified to include measurements of required metrics, while static information such as authors, dataset metadata, and device properties should be collected and inserted into the dataset.

Such a dataset can be inserted in a Neo4j instance in the same manner as our example localization \ac{m3kg}, discussed in Section~\ref{subsec:loc_kg} while the entire process is documented in \ac{m3kg} Github repository\footnotemark[5]. The practical implementation requires deploying a Neo4j graph database instance, which can be accomplished using Docker containerization. Prior to knowledge graph construction, \ac{json} files should be validated against the model card schema using the provided validation utility to ensure the correct structure.

The transformation process from \ac{json} metadata to graph entities involves creation of nodes and relationship while following the structure defined in the ontology.

Once deployed, this domain-specific metadata knowledge graph enables use cases including automated model selection based on specific constraints such as energy consumption as presented in Listing~\ref{listing:cypher} in Section~\ref{subsec:queryingKG}.

As new ML models are trained and new metadata information is collected, this data can be inserted into the deployed Neo4j instance following the same process outlined in the preceding paragraphs to maintain consistent ontology structure. 
The transformation from \ac{json} files to graph entities remains unchanged. 
This continuous update of the knowledge graph enables selection of newly added models across an expanding collection of ML model metadata.

\section{Conclusion}
\label{sec:conclusion}
This paper presents a solution for structured representation of ML model metadata through a knowledge graph approach. Our proposed \textit{Machine Readable ML Model Metadata} addresses the critical gap in machine-readable model documentation by establishing a standardized ontology that systematically captures relationships between models, datasets, architectures, and performance metrics. By implementing this ontology in Neo4j and showing an example localization \ac{m3kg} with 22 localization models trained on 4 datasets, we demonstrated the practical utility of our approach, enabling efficient querying of models based on multiple criteria including environmental impact metrics. The standardized \ac{json} schema we developed enables consistent metadata collection across diverse ML projects, simplifying integration with existing workflows. 


In our future work, we will show and implement the discussed use cases, such as automated selection of ML models for the example localization \ac{m3kg} and extend the domain to ML models beyond network systems.

\begin{acks}
This work was supported by the Slovenian Research Agency (P2-0016) and the European Commission NANCY project (No. 101096456).
We would like to acknowledge the SensorLab team for developing the localization models, and Gregor Cerar and Tim Strnad for developing an automated training pipeline. 
\end{acks}

\bibliographystyle{ACM-Reference-Format}
\bibliography{references, mendeley-ref}

\end{document}